%% file: paper.tex
\documentclass[twocolumn]{svjour3}

\usepackage{times}
\usepackage{epsfig}
\usepackage{graphicx}
\usepackage{amsmath}
\usepackage{amssymb}
\usepackage{comment}
\usepackage{defs}

\DeclareRobustCommand\onedot{\futurelet\@let@token\@onedot}
\def\@onedot{\ifx\@let@token.\else.\null\fi\xspace}
 
\def\ie{\emph{i.e}\onedot}

\def\etal{\emph{et al}\onedot\xspace}

\title{Boosting in Location Space}
\author{Damian Eads \and David Helmbold \and Ed Rosten}
\institute{Department of Engineering, Trumpington Street, Cambridge, England CB2 1PZ \and Department of Computer Science, University of California, 1156 High Street, Santa Cruz, CA 95060 \and Department of Engineering, Trumpington Street, Cambridge, England CB2 1PZ}
\begin{document}
\maketitle
\subclass{Machine Vision and Applications}
\begin{abstract}
The goal of object detection is to find objects in an image.  An object
detector accepts an image and produces a list of locations as $(x,y)$
pairs.  Here we introduce a new concept: {\bf location-based
boosting}.  Location-based boosting differs from previous boosting
algorithms because it optimizes a new spatial loss function to combine
object detectors, each of which may have marginal performance, into a
single, more accurate object detector. A structured representation of
object locations as a list of $(x,y)$ pairs is a more natural domain
for object detection than the spatially unstructured representation
produced by classifiers. Furthermore, this formulation allows us to
take advantage of the intuition that large areas of the background are
uninteresting and it is not worth expending computational effort on
them. This results in a more scalable algorithm because it does not
need to take measures to prevent the background data from swamping the
foreground data such as subsampling or applying an ad-hoc weighting to
the pixels. We first present the theory of location-based boosting,
and then motivate it with empirical results on a challenging data set.
\end{abstract}

\input{introduction}

\input{background}

\input{algorithm}

\input{practical}

\input{conclusions}

\bibliographystyle{spmpsci}
\bibliography{egbib}

\end{document}

%% file: introduction.tex
\section{Introduction}

Machine learning approaches to object detection often recast the
problem of localizing objects as a classification problem.
This is convenient because it allows standard machine learning techniques to
be used,  but  does not exploit the full structure of the
problem. 
Such approaches usually learn classifiers that, when applied to an image patch,
predict whether or not the patch contains an object of interest.
To find the objects in an image, one first applies 
the learned classifier to all possible sub-windows and then arbitrates
overlapping detections.  
An especially popular approach combines weak image
classifiers of local image patches using a boosting algorithm such as
AdaBoost\cite{violajones01}.
Although windows with strong detections provide bounding box estimates,
Blaschko and Lampert~\cite{blaschko08} noted that the training optimizes the classifier for detection rather than localization.

In contrast, we consider the problem of directly localizing the objects by finding the center coordinates of (usually many) small 
objects in the image (see Figure~\ref{f:azim}). 
Since the objects can be as small as a few dozen pixels, this problem has a distinctly different 
character than the well-studied large object detection problem (e.g.~the PASCAL
dataset~\cite{everingham2006pvo})
and requires different techniques.
Sliding window/bounding box techniques can be handicapped by the lack of surrounding context
\cite{divvala09esod}, and
expanding the windows to provide this context risks reducing the localization accuracy and collapsing 
nearby objects into a single detection.
The individual objects are already very small, so parts-based detection is unlikely to be helpful (although
a small object detector could be useful for detecting and locating parts within a large object detector).
Pixel-based methods,
which are often equivalent to sliding window methods, 
have had some success despite difficulties with spurious detections and arbitrating between nearby objects.
They also illustrate a shortcoming of classification based methods. A classifier
finding all of the pixels on half of the objects will have the same score as one finding
half of the pixels on all of the objects. The latter, however, is a much better object detector.

The main contribution of this paper is a new boosting-based methodology for object detection that
works directly in location space rather than going through a classification step.
In this work, a \emph{location-based} object detector takes as input an image
and produces a list of $(x,y)$ pairs where each
pair in the list is the predicted center of an object.
Instead of focusing on finding new features or new ways of casting
object detection as classification, 
we present \underline{L}ocation-\underline{B}ased \underline{Boost}ing (LB-Boost), an algorithm
specifically tailored to location-based object detectors.  
This algorithm treats images in their entirety
rather than using windows as its basic units.
LB-Boost is
different from other boosting schemes because it combines location
based detectors, not binary classifiers.  
In particular, LB-Boost
learns a weighted combination (an {\em ensemble}) of location-based object detectors and
produces a list of $(x,y)$ locations.
Since both the ensemble and its components are location-based object detectors,
our approach is composable and can be viewed
as a meta-object detection methodology.

At each iteration, LB-Boost incorporates a new object detector into its  \emph{ensemble}. 
The new object detector's predicted locations provide evidence of object centers
while an absence of objects may be suggested in those areas away from the predicted locations.
Following the InfoBoost algorithm~\cite{aslam00},
LB-Boost uses two different weights for these different kinds of evidence; this decouples and simplifies 
the weight optimization while speeding convergence.
The \emph{master detector} for the ensemble
uses the weighted combination of the 
ensemble members' evidence to produce a final list of predicted locations.

LB-Boost uses a new spatially-motivated loss function 
to drive the boosting process.
Previous boosting-based sliding window and pixel classifiers
use a smooth and strictly positive loss function (like AdaBoost's
exponential loss) summed
over both the foreground and background.  
In contrast, the loss function we propose does not penalize background
locations unless an object detector in the ensemble generates a false
detection at that location.  
Therefore LB-Boost effectively ignores 
the large amounts of uninteresting background
and focuses directly on the object locations.
This emphasis on the foreground makes training on entire images feasible, 
eliminating the need for subsampling windows from the background.
As in standard boosting methods, the loss of the master detector
is positive and decreases every iteration.

%% file: background.tex
\section{Background and Previous Work}

In order for machine vision systems to understand an image or a visual
environment, they must be able to find objects. Object detection is
thus an important area of research but the problem is ill-posed,
making a unified and guarantee-based approach elusive. In recent
years, the community has been headed toward formalizing different object
detection problems, which will facilitate the study of more rigorous
approaches to learning in vision systems.

Our techniques are different form the typical {\em sliding window} approach 
to object detection~\cite{violajones01,ferrari08}
where an image classifier is applied to every
sub-window of an image in order to quantify how likely it is that there 
is an object within the window.
The image patch
within the window, either processed or unprocessed, is then taken as a feature
vector. This allows standard machine learning methods to be used to determine
the presence or absence of an object within the window.  A good sliding window
approach must generate confidences meaningful to the vision problem domain at
hand and must carefully arbitrate between nearby detections. 
Recently, Alexe~\etal~\cite{alexe10wiao} identify
necessary characteristics of sliding window confidence measures, and propose a
new superpixel straddling cue, which shows strong performance on the PASCAL
dataset~\cite{everingham2006pvo}.
Expanding the window to include context from outside the object's bounding box 
can improve detection accuracy at the expense of localization
(see the discussion and references in~\cite{blaschko09} for some examples).
Our location-based methodology is holistic and does not restrict the context 
that can be used by the ensemble members.

There are several overlapping categories of object detection methods.
{\em Parts-based} models, as the name implies, break down an object
into constituent parts to make predictions about the
whole~~\cite{fergus03ocr,agarwal2004ldo}. Some model the relative
positions of each part of an object while others predict based on just
the presence or absence of parts~\cite{heisele2003frc,zhang2007lfa}.
Heisele, \etal~\cite{heisele2003frc} train a two-level hierarchy
of support vector machines: the first level of SVMs finds the presence
of parts, and these outputs are fed into a master SVM to determine the
presence of an object. 
Although some ensemble members might be similar to parts detectors,
our ensemble members are optimized for their discrimination and localization 
benefits rather than being trained on particular parts of the objects.

In this paper we do not consider {\em segmentation}  which involves fully separating the object(s) of interest from other
objects and background using either polygons or pixel
classifiers~\cite{shotton2008stf}.  
However, we note that some object detection and localization approaches do exploit
segmentation or contour information (for example,~\cite{ferrari08,opelt2006bfm,fulkerson09}).



A large number of object detectors use {\em interest point detectors} to find
salient, repeatable, and discriminative points in the image as a first
step~\cite{dorko03,agarwal2004ldo}. Feature descriptor vectors are
often computed from these interest points. 
The system described here uses
a stochastic grammar to randomly generate image features for the ensemble members,
although 
 ensemble members derived from
interest points and their feature descriptor vectors may be a fruitful direction
for future work.  


Boosting is a powerful general-purpose method for 
creating an accurate ensemble from 
easily learnable weak
classifiers that are only slightly better than random guessing.
AdaBoost~\cite{freund96experiments} was one of the
earliest and most successful boosting algorithms, in part due to its
simplicity and good performance.
The boosted cascade of Viola and Jones~\cite{violajones01} was one of the first applications
of boosting for object detection.
%
Instead of using boosting to classify windows, we 
propose a new variant of boosting that is
specially tailored to the problem of object
detection. 
In particular, the learned master detector outputs a list of predicted object centers
rather than classifying windows as to whether or not they contain objects.
Aslam's InfoBoost algorithm~\cite{aslam00} is like AdaBoost, but  uses two weights for each weak classifier in the ensemble:
one for the weak classifier's positive predictions and one for its negative predictions.
This two-weight approach allows the ensemble to better exploit the information provided by marginal predictors.
In our application the two-weight approach also has the benefit of decomposing (and thus simplifying) the 
optimization of the weights.

The Beamer system of Eads~\etal~uses boosting to build a pixel-level classifier
for small objects \cite{eads2009bmvc}. 
The pixel based approach requires pixel-level markup on the training set instead of simply object centers, and
we show improved results with our object-level system on their dataset in Section~\ref{s:experiments}.

Blashcko and Lampert's work~\cite{blaschko08} may be the most similar to ours in spirit.
Rather than simply classifying windows, they use structural SVMs to directly output object locations.
From each image their technique produces a single best bounding box and a bit indicating if an object
is believed to be present.   
Our methodology is based on boosting rather than SVMs, and detects multiple object centers
as opposed to a single object's bounding box.

%% file: algorithm.tex
\section{Learning Algorithm}

Location-based boosting is a new framework for learning a weighted ensemble of
object detectors.  
As in the functional gradient descent view of
boosting~\cite{friedman01,mason00}, the learning process iteratively attempts to
minimize a loss function over the labeled training sample.  
Since we are learning at the object rather than pixel level, each training
image is labeled with a list of $(x,y)$ pairs indicating the object centers,
rather than detailed object delineations.
In each iteration a
promising weak object detector is added into the {\em master detector}.  Most
boosting-based object detection systems use weak classifiers that predict the
presence or absence of an object at either the pixel or window level.  In
contrast, this work uses a different type of weak hypotheses that predicts a set
of object centers in the image and the boosting process 
minimizes a spatial loss function. 
Our loss function (Section~\ref{s:optimize}) encourages maximizing the predictions
at the given object locations while keeping the predictions on the background
areas below zero. 
This allows large areas of uninteresting background to be
efficiently ignored.  
The form of the loss function is chosen so that the optimization is tractable.

\input{weak}
\input{hos}
\input{optimizing}

\input{theta}

%% file: weak.tex
\subsection{Object Detectors and  `Objectness'}

An object detector may be simple, like finding the large local maxima of 
an image processing operator. It can also be complex, such as
the output of an intricate object detection algorithm. 
When a (potentially expensive) object detector attaches confidences
to its predicted locations, one gets a whole family of object detectors by
considering different confidence thresholds.

A \emph{confidence-rated object detector} is given an input image
(or set of images) and
generates a set of confidence-rated locations
$h = \{((x_1,y_1), c_1)$,$((x_2,y_2), c_2)$, $\ldots$ , $((x_n,y_n), c_n)$$\}$
where $(x_i,y_i)$ is the $i$'th predicted location in the image and $c_i$ is the confidence of the prediction.
The confidences in this list define an ordering over the detections,
and we use $h(\theta)$ for the list $h$ filtered at confidence threshold $\theta$:
\begin{equation}
    h(\theta)=\{(x,y) : ((x,y),c) \in h \tand c \geq \theta\}.
\end{equation}
Each confidence-rated object detector will be coupled with an optimized threshold when it is added to the ensemble,
and the master detector uses the filtered list of locations to make its predictions.
Since confidence-rated object detectors can have radically different confidence scales, 
it is difficult for the ensemble to make post-filtering use of the confidence information.
During the training state, this filtering approach allows the efficient consideration of many candidate object detecters 
derived from the same (possible expensive) confidence-rated predictor.

It is unlikely that the
locations predicted by an object detector will precisely align with the centers of the desired objects,
but they do provide a kind of evidence that an object center is nearby.
We use a non-negative correlation function $C(\xv,\yv)$ to measure the evidence that an object 
is at location $\xv$ given by the predicted location $\yv$
(we use $\xv$ and $\yv$, as well as $(x,y)$ to denote locations in the image).
There are several alternatives for the function $C$, which one is most appropriate may depend
on the particular problem at hand. 
A simple alternative is to use a flat disk:
\[
  	C(\xv, \yv) = \left\{
   		\begin{array}{cl} 
			1 & \text{if}\ || \xv - \yv ||_2 < r\\
			0 & \text{otherwise,}
   		\end{array}
	\right.
\]
where $r$ is the correlation radius.
Other natural choices include  $C(\xv,\yv) = \max \{ 0, 1-d(\xv,\yv)\}$ or $C(\xv,\yv) = \max \{ 0, 1-d(\xv,\yv)^2\}$
(where $d$ is a distance function).
A whole family of alternatives view the object location $\xv$ and/or the predicted location $\yv$ as corrupted with a little bit of noise, 
and $C(\xv,\yv)$ is the convolution of these noise processes.
For efficiency reasons we require that $C(\xv,\yv)$ is non-zero only when $\xv$ and $\yv$ are close (allowing the
bulk of the background to be ignored).
Without loss of generality we assume that the correlation is scaled so $C(\xv,\yv)\leq 1$.

We use $f(\xv; h(\theta))$ for the \emph{evidence}\footnote{We use ``evidence'' informally rather than in its statistical sense.}
given by object detector $h(\theta)$ that an object is at an arbitrary location $\xv$.
One can think of $f(\xv; h(\theta))$ as the ``objectness'' assigned to $\xv$ by $h(\theta)$
(unlike the generic  ``objectness" coined by Alexe~\etal~\cite{alexe10wiao}, here ``objectness" is a 
location's similarity to the particular object class of interest).
Although it is natural to set $f(\xv: h(\theta)) = \sum_{\yv\in h(\theta)} C(\xv,\yv)$,
It will be convenient for the evidence to lie in $[0,1]$, and the sum can be greater when
$h(\theta)$ contains several nearby locations.
Therefore we define $f(\xv;h(\theta))$ as   
as either capped:
\begin{align}
  f(\xv; \theta)&=\min \left\{1, \sum_{\vv \in h(\theta)}{ C(\xv,\vv) }\right\}
  \label{eqn:cap}
\intertext{or uniquely assigned from the closest detection:}
  f(\xv; \theta)&=\max_{\vv \in h(\theta)}{ C(\xv,\vv) }.
  \label{eqn:unique}
\end{align}
Our experiments indicated little difference between these two choices.

%

\begin{figure*}
\begin{tabular}{ccc}
\fbox{\includegraphics[width=4.7cm]{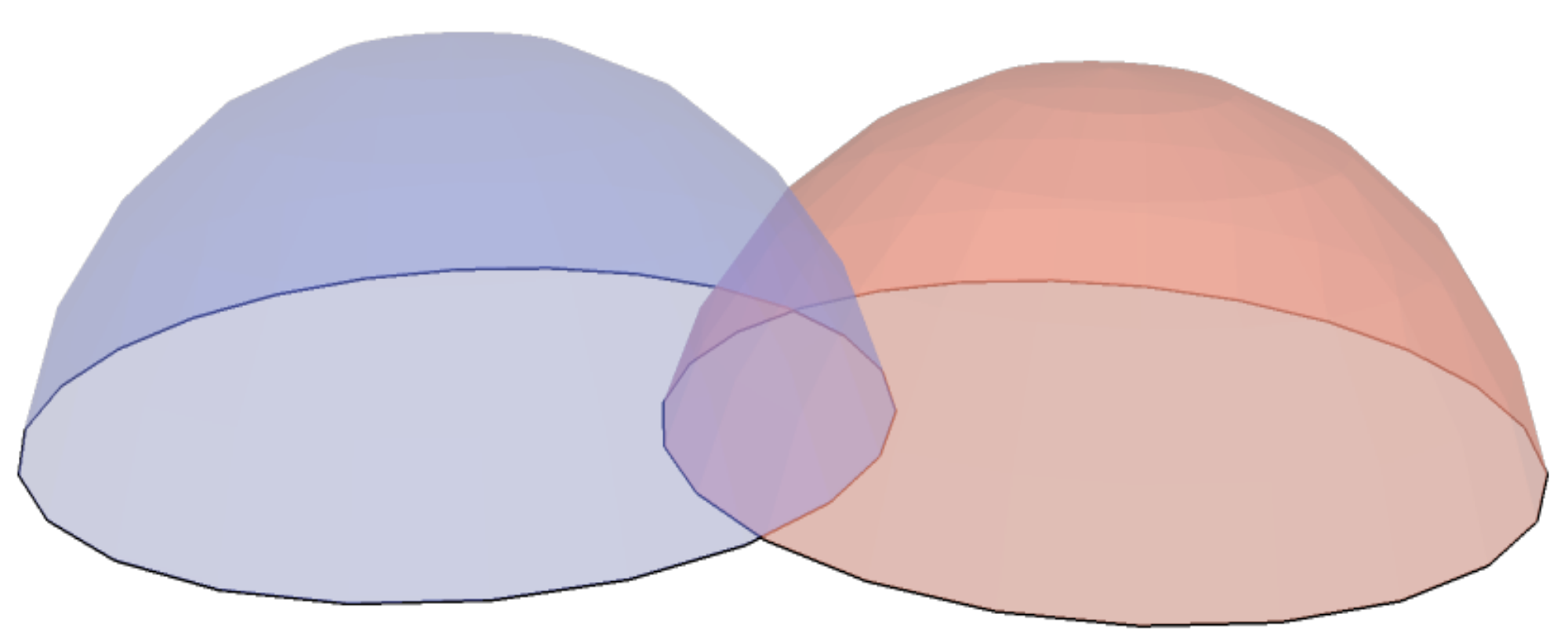}}&
\fbox{\includegraphics[width=4.7cm]{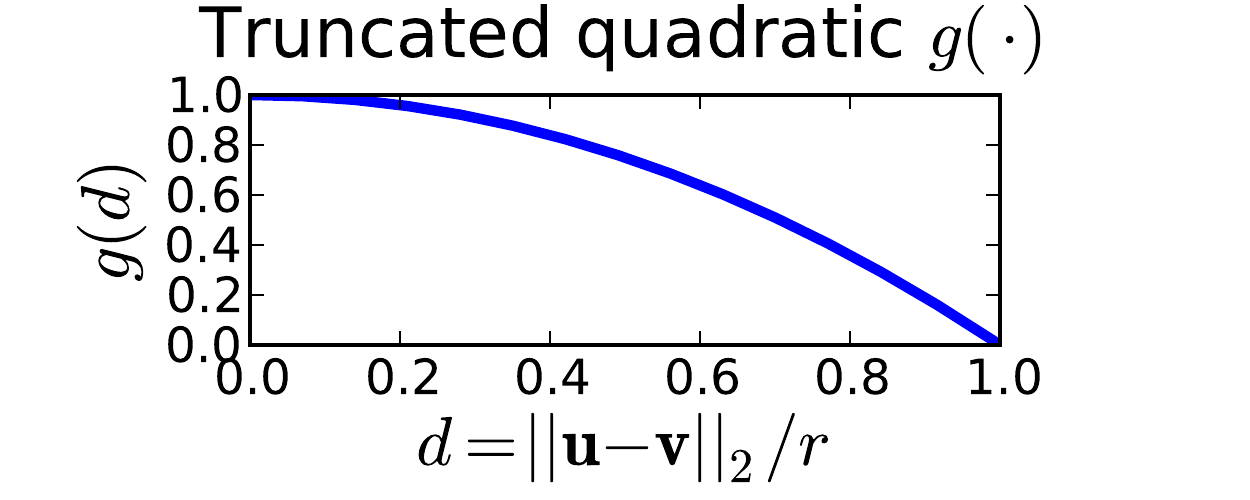}}&
\fbox{\includegraphics[width=4.7cm]{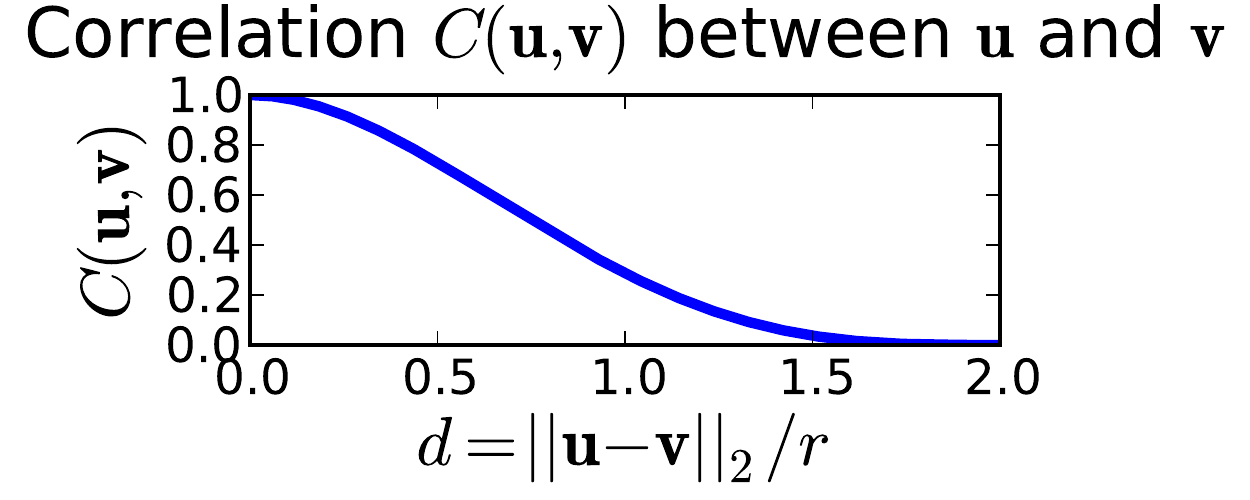}}\\
(a)&(b)&(c)
\end{tabular}
\caption{Plots of (a) two overlapping quadratic bumps with centers $\xv$ and $\vv$, (b) the truncated quadratic kernel $g$ as a function of  distance $d=||\xv-\vv||_2/r$, and (c) the correlation $C$.}
\label{f:kernels}
\end{figure*}

%% file: hos.tex
\subsection{Hit-or-Shift filtering and the master hypothesis}

If weak hypotheses could only add objectness to the master hypothesis, then the
objectness due to a false positive location would be  impossible to remove.  However,
a lack of detections \emph{may} be interpreted as evidence of absence of
objects.  To handle this, we process the weak hypothesis to reduce the
objectness in those image locations which are uncorrelated with the locations
predicted by the hypothesis, $h(\theta)$.

The negative influence is achieved using the HoS (Hit or Shift) filter.
The filter is designed to make optimization tractable.
A {\em hit} is a location with positive objectness which exerts positive influence on a master hypothesis.
In contrast, the shift is a location with no objectness which exerts negative influence.
Recall the definition of $f(\xv; \theta) = \max_{\vv\in h(\theta)} C(\xv, \vv)$.
Given a location $\xv$, a HoS weak hypothesis predicts the positive quantity $\alpha f(\xv)$ 
only when $f(\xv;\theta)$ is positive and otherwise predicts  $-s$:
\begin{equation}
  f'(\xv) = \begin{cases} \alpha f(\xv;\theta) & \text{if $f(\xv; \theta) > 0$,}  \\
                                            -s & \text{if $f(\xv; \theta)=0$}.
                     \end{cases}
\end{equation}
The HoS formulation is useful because it allows the optimization of $\alpha$ and $s$
to be performed as two independent one dimensional searches rather than a two
dimensional optimization.

To avoid cluttering the notation, we keep the HoS hypothesis parameters $\alpha$, $s$, and $\theta$ implicit.
We assume that the detectors are positively correlated with objects, 
implying $s \ge 0$ and $\alpha \ge 0$;  this simplifies the optimization.

Together, the two parameters $\alpha$ and $s$ allow us to create a weighted sum
of HoS detectors. This is analogous to AdaBoost, which creates a weighted sum
of weak hypotheses, where each hypothesis has only a single weight.  Using two
weights can make better use of hypotheses, for example, a weak hypothesis that
has few false positives but finds only some of the objects can contribute a
large $\alpha$ and a small $s$ even though its overall accuracy may be poor.

The master hypothesis uses an ensemble of HoS filtered hypotheses, and we
subscript $h$, $f$, and $f'$ to indicate the ensemble members.  The master
hypothesis from boosting iteration $t$ is the cumulative {\em objectness} of
the $t$ HoS weak hypotheses in the ensemble:
\begin{equation}
  H_t(\xv)=\sum_{i=1}^{t}{f'_i(\xv)}.
\end{equation}
$H_t(\xv)$ predicts objects where $H_t(\xv) > 0$, and background otherwise. It
can be turned into a confidence rated detector producing the list 
$\{((x_1, y_1),c_1), \cdots ((x_n, y_n), c_n)\}$ where $(x_i, y_i)$ are positive
local maxima of $H_t$ and  the confidence, $c_i = H_t((x_i, y_i))$, are the height of
the maxima.

%% file: optimizing.tex
\subsection{Optimizing the weight and shift parameters} \label{s:optimize}

\def\Exp#1{e^{#1}}

Optimizing the performance of $H$ requires an objective function.
The objective function is defined over the training set so that the position of objects
is known.
To make optimization tractable, our
objective is split into two parts, the loss on objects and the loss on background.
We define the loss on objects to be:
\begin{equation}
 \Lfg = \sum_{\xv \in \mbox{obj}} {\Exp{-H_t(\xv)}},\label{e:fgloss}
\end{equation}
where `obj' is the set of object locations.
$\Lfg$ generates a very large loss if strong background is predicted where an 
object exists, whereas the loss rapidly shrinks if an object is predicted in the right
place.
We define the 
loss outside the objects of interest, \ie on all background pixels, `bg', as:
\newcommand{\set}[1]{{\left\{ #1 \right\}}}
\begin{equation} \label{e:bgloss}
\Lbg = b \sum_{\xv \in \mbox{bg}} {\max \set{0, \Exp{H_t(\xv)}-1}} ,
\end{equation}
where $b = \frac{|\text{obj}|}{|\text{bg}|}$ is the background discount which sets the trade-off between false
positives and false negatives. This generates a large loss if an object is
predicted in the background, but generates absoloutely no loss on background
correctly predicted. 
Therefore $\Lbg$ allows vast swaths of easily learned background to 
be efficiently ignored because the loss on those regions never rises above zero.
The total loss is $\Lfg + \Lbg$.  This loss function puts more emphasis on
correctly learning object concepts and less on learning unimportant object
background than the exponential loss associated with AdaBoost.

Note that the training set, $\text{obj}\cup\text{bg}$ is not necessarily the
entire image, as pixels near to `obj' are not background pixels. No
loss is computed on these intermediate pixels. These intermediate ``don't care''
regions may be taken to be a disc shaped region around each object location.

At each iteration $t$, the new master hypothesis $H_t$ is created by selecting a new weak hypotheses $h_t$
and adding its objectness function $f'_t$ to the previous hypothesis $H_{t-1}$.
To optimize the parameters of a candidate hypothesis $h$ it is convenient to
re-partition the loss by partitioning the training set. The partitioning
allows us to separate the loss into components which depend only on $s$ or
only on $\alpha$. This allows up to optimize $s$ and $\alpha$ independently.

The pixels in `obj' for which $f_t$ is positive
are:
\begin{align}
	\fgpos&=\{\xv \in \fg : f_t(\xv)>0\},\\
\intertext{likewise, we define the other partitions as:}
    \bgpos&=\{\xv \in \bg : f_t(\xv)>0\},\\
	\fgzero&=\{\xv \in \fg : f_t(\xv)=0\},\\
    \bgzero&=\{\xv \in \bg : f_t(\xv)=0\}.
\end{align}

For brevity, we write $H(\xv)\equiv H_{t-1}(\xv)$ and $f(x) \equiv f_t(x)$.  The total loss of $H_{t}$
can be written as a sum of two loss functions, the {\em alpha loss}, which
depends only on $\alpha$ and the {\em shift loss} which depends only on $s$.

The alpha loss can be expressed as:
\begin{equation}
\begin{split}
  \Lal_t=&\sum_{\xv \in \fgpos} \Exp{-\alpha f(\xv) - H(\xv)}+\\
         &\ \ \ \ \ b\sum_{\xv \in \bgpos}{\max\{0,\Exp{\alpha f(\xv)+ H(\xv)}-1\}} ,
  \label{e:alphaloss}
\end{split}
\end{equation}
and the {\em shift loss} which depends only on $s$ is:
\begin{equation}
\begin{split}
  \Ls_t=&\sum_{\xv \in \fgzero} \Exp{-H(\xv)} \Exp{s} + 
        b \sum_{\xv \in \bgzero} \max\set{0,\Exp{H(\xv)} \Exp{-s}-1 }.
  \label{e:shiftloss}
\end{split}
\end{equation}
$\Ls_t$ is a sum of non-negative convex functions in $s$, and thus is convex in $s$. 
When the set $\fgzero$ is empty (i.e. no false negatives) then the shift loss
$\Ls_t$ can be made zero by setting $s$ to $\max_{\xv \in \bgzero}H(\xv)$.
Otherwise, the derivative of 
$\Ls_t$ is piecewise continuous, with discontinuities when $s=H(\xv)$ for
some $\xv \in \bgzero$.

Since the first sum in Equation~\ref{e:shiftloss} is independent of $s$, 
we define $V = \sum_{\xv \in \fgzero} \Exp{-H(\xv)}$ for convenience.
$\Ls_t$ can be be rewritten as:
\begin{equation}
\Ls_t = V\Exp{s} + b\sum_{i=1}^nm_i \max\set{0, \Exp{k_i}\Exp{-s} -1},
\end{equation}
where $k_i$ are distinct values such that \mbox{$k_1 < \cdots < k_n$, $m_i = |\set{ \xv \in \bgzero, H(\xv) =
k_i}|$} and $m_i \ge 1$.  The index $j \le i$ is the index where
$\Exp{k_{i} - s} \le 1$.
By sorting the background values in this manner, we can easily
split the values in to two groups: those for which the max operator
returns zero and those for which it does not. Only the latter
contribute to the sum.
A locally correct equation for $\Ls_t$ is:
\begin{equation}
\Ls_t(j) = V\Exp{s} + \Exp{-s}b\sum_{i=j}^n m_i\Exp{k_i} - b(n-j+1)
\label{eq:locals}
\end{equation}
where $\Ls_t(j)$ is valid in the range $k_{j-1} \le s \le k_j$. $\Ls_t$ can be easily
optimized in closed form, making the optimization of $s$ very efficient as
$\Ls_t(j)$ is naturally computed incrementally from $\Ls_t(j+1)$. 
By setting the derivative to zero, we get the unconstrained minimizer of the locally correct 
loss:
\begin{align}
{\hat{s}^{*}}(j) &= \frac{1}{2}\ln\frac{\sum_{i=j}^nm_i\Exp{y_i}}{V}\\
\intertext{but since $s(j)$ is constrained, the minimizer
is:}
s^*(j) &= \max \left\{k_{j-1}, \left\{ \min {\hat{s}^{*}}(j), k_j \right\}\right\}.
\end{align}


Efficiently optimizing $\alpha$ is similar, but trickier.
We begin by rewriting the alpha loss to remove the maximum,
\begin{equation}
\begin{split}
\Lal_t = &\sum_{\xv  \in \fgpos} \Exp{-H(\xv)} \Exp{-{\alpha f(\xv)}} + \\
 & b \!\!\!\! \sum_{\xv \in \bgpos: \alpha > \frac{- H(\xv)}{ f(\xv)}} \!\!\!\!
	\left(\Exp{H(\xv)}\Exp{\alpha f(\xv)} -1 \right).
\end{split}
\end{equation}
We assume $0 \leq f(\xv) \leq 1$ which can be enforced by either
capping (Equation~\ref{eqn:cap}) or uniqueness (Equation~\ref{eqn:unique}).
Even with this assumption, the 
$\exp(\alpha f(\xv))$ terms are problematic.

Similarly to $\Ls$ we define $z_1 < \cdots < z_n$ to be distinct values such that
$A_i = \set{\xv \in \bgpos, \frac{-H(\xv)}{f(\xv)} = z_i}$ and $A_i\neq
\emptyset$. Additionally, we define $z_0=0$ and $A_0 = \set{\xv \in \bgpos,
H(\xv)
> 0}$. $A_i$ therefore partitions the set $\bgpos$.

We can now write a locally correct $\Lal_t(j) = \Lal_t$ where $y_j < \alpha \leq
y_{j+1}$:
\begin{equation}
\begin{split}
\Lal_t(j) = &\sum_{\xv \in \fgpos}\Exp{-H(\xv)-\alpha f(\xv)} +  
            \sum_{i=0}^{j}\sum_{\xv \in A_i} \left[ \Exp{H(\xv)+\alpha f(\xv)} - 1\right]
\end{split}
\end{equation}

We overestimate $\Lal_t$  as $\hat{\Lal_t}$ by approximating $\exp(\pm \alpha f(\xv))$ with 
$1-f(\xv) + f(\xv) \exp(\pm \alpha)$ and then minimize this overestimate. 
Like the shift loss, the overestimate is convex and its 
derivative:
\begin{equation}
\frac{\partial\hat{\Lal_t}(j)}{\partial\alpha} = \Exp{-\alpha}\sum_{\xv \in \fgpos}\Exp{-H(\xv)}f(\xv) +  
            \Exp{\alpha}\sum_{i=0}^{j}\sum_{\xv \in A_i} \Exp{H(\xv)}f(\xv)
\end{equation}
is piecewise continuous.  
The value of $\alpha$ that minimizes the overestimate is either
a point where the derivative is discontinuous or  
the solution to setting the derivative in one of its piecewise continuous regions to zero.
Since $\Lal_t$ equals the overestimate when $\alpha=0$, any $\alpha>0$
minimizing the overestimate also reduces the alpha loss $\Lal_t$.

Note that if flat kernels are used ($f(\xv) \in \set{0,1}$), then an
approximation is not required and the locally correct loss is:
\begin{equation}
\Lal_t(j) = \Exp{-\alpha}\sum_{\xv \in \fgpos}\Exp{-H(\xv)} +
b\sum_{\makebox[0pt][c]{$\scriptstyle \xv \in \bgpos ; -H(\xv) >
q_j$}} \Exp{\alpha}\Exp{H(\xv)} -1
\end{equation}
where $q_1<\cdots<q_n$ are the distinct positive values of $H(\xv)$. This 
equation is strictly convex and can be optimized in a manner similar
to $\Ls$.

%% file: theta.tex
\subsection{Finding the best $\theta$}
The previous discussion assumed that the threshold $\theta$ for filtering
the locations in $h$ had already been chosen.
However, the effectiveness of a weak hypothesis often depends on choosing an
appropriate threshold, and naively searching over $(\theta,\alpha,s)$ triples 
is computationally expensive even though the searches for $\alpha$ and $s$ can
be decoupled.  
Fortunately, there is structure associated with $\theta$ that enables some efficiencies.

When threshold $\theta$ is above the highest confidence, $h(\theta)$ contains no
locations.  
As $\theta$ drops, locations are added to $h(\theta)$ in confidence order.
As locations are added to $h(\theta)$ we incrementally update 
the $\fgpos$, $\bgpos$, $\fgzero$, and $\bgzero$ sets
as well as the local approximations and related sums needed to 
find the $s$ and $\alpha$ parameters.
This exploitation of previous computation greatly speeds the optimization 
of $s$ and $\alpha$ for the new threshold and allows an exhaustive search over
$\theta$ thresholds.
Therefore the best parameters found provably minimise the 
empirical loss estimate $\Ls+\hat{\Lal}$  within the parameterized family of weak hypotheses associated with $h$.

%% file: practical.tex
\section{Practical application of location-based boosting to detecting small objects}

Location-based boosting is an abstract algorithm; it creates an ensemble using a
source of confidence-rated detectors.  One way to create detectors is to take
a feature intensity image and convert its local maxima into predicted locations
with the height of the local maxima used as the confidence.

In each boosting iteration we generate several (typically 100) random features
and optimize their $\theta$, $\alpha$, and $s$ parameters.  The resulting
hypothesis that results in the minimum loss is incorporated into the master
hypothesis.

The labeled data is split into three partitions. The first partition
is to train the HoS ensemble. The second partition, known as
validation, is used to train other parameters. The third partition,
known as testing, is used to evaluate the performance of the
algorithm.

\paragraph{Easy labeling} One benefit of location-based boosting is that only object centers
need to be annotated.  This simplifies the labeling of training data.
\vspace{-3ex}

\paragraph{``Don't Care'' regions} When an HoS
hypothesis predicts an $(x,y)$ location, the nearby pixels also accumulate
objectness.
However, locations close
to an object center are likely to be part of the object, and thus
should not be treated as background. 
When training, we create a disc-shaped \emph{``don't care'' region} of radius $\rho$ 
around each object location, and the background loss calculations
omits the locations within these ``don't care'' regions.
\vspace{-3ex}

\begin{figure*}
\begin{center}
\begin{tabular}{cccc}
\fbox{\includegraphics[height=2.5cm]{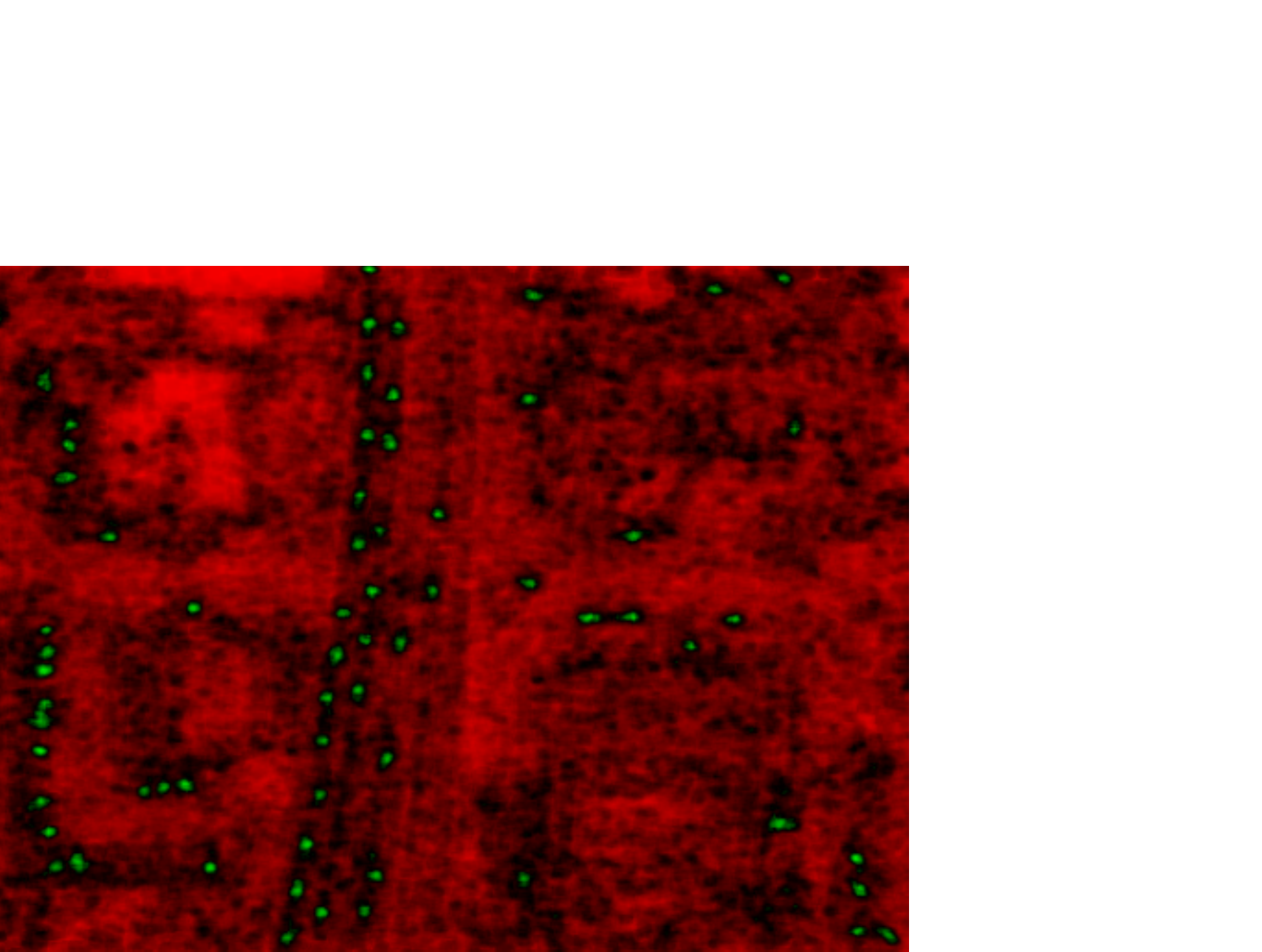}}&
\fbox{\includegraphics[height=2.5cm]{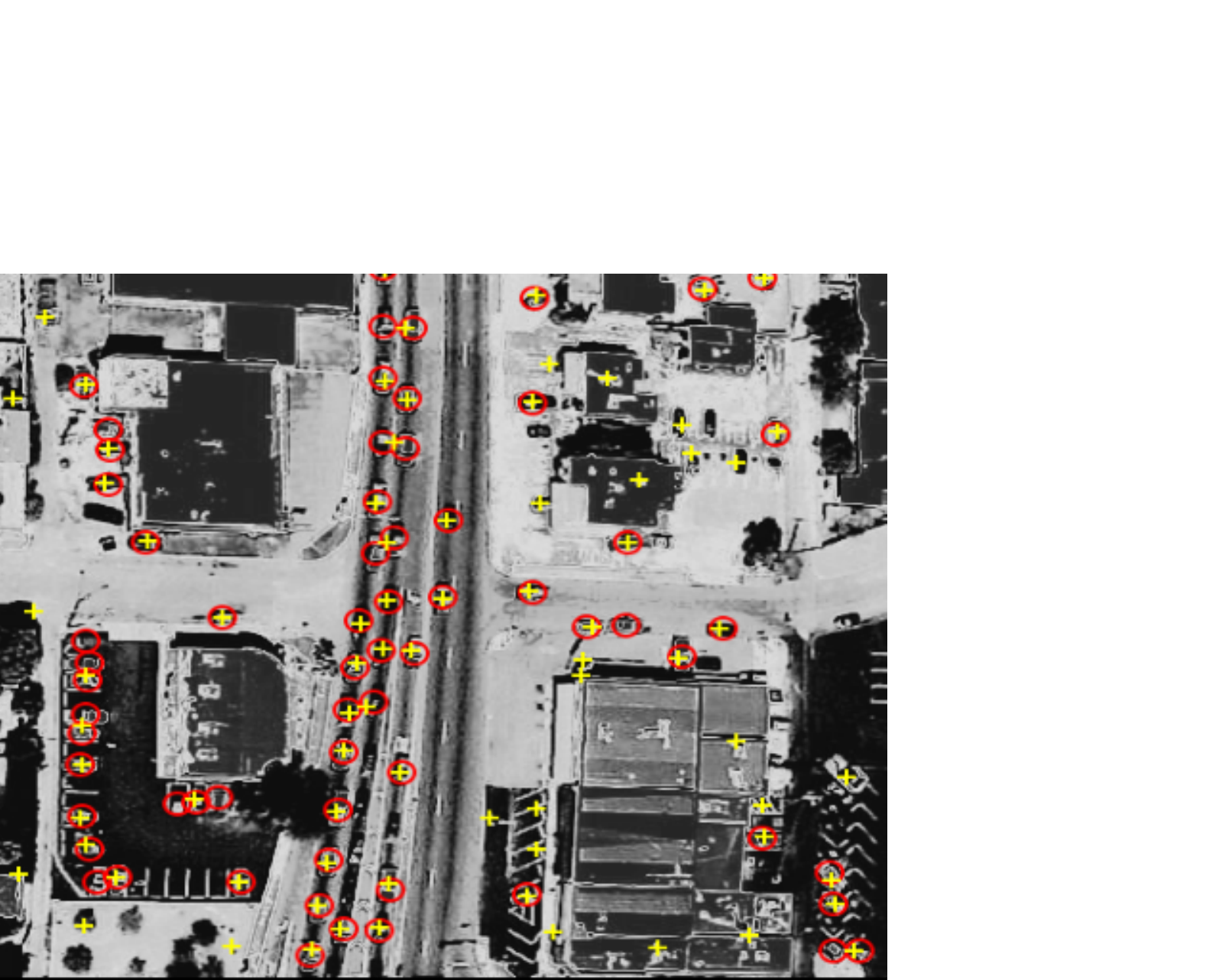}}&
\fbox{\includegraphics[height=2.5cm]{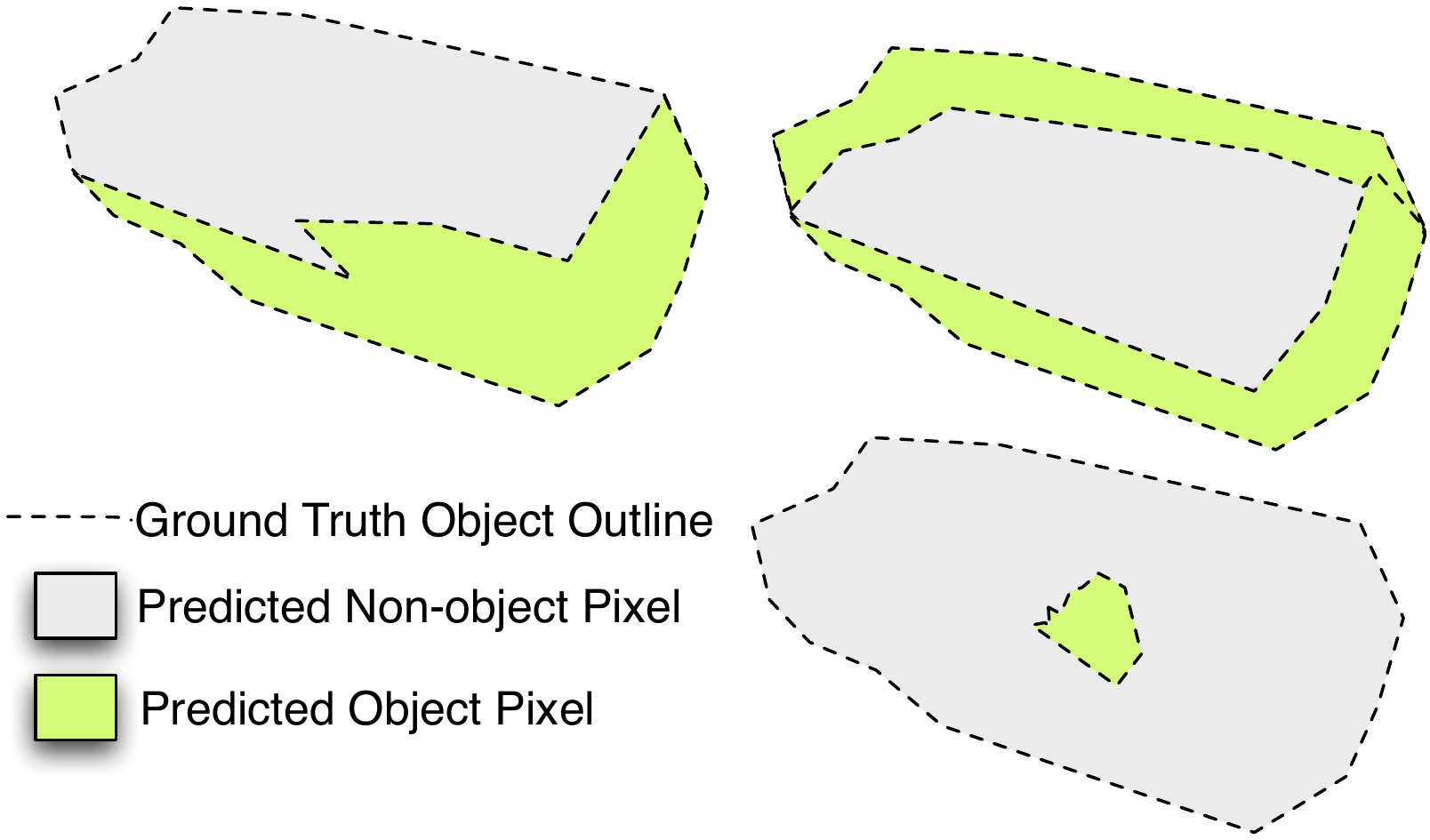}}&
\fbox{\includegraphics[height=2.5cm]{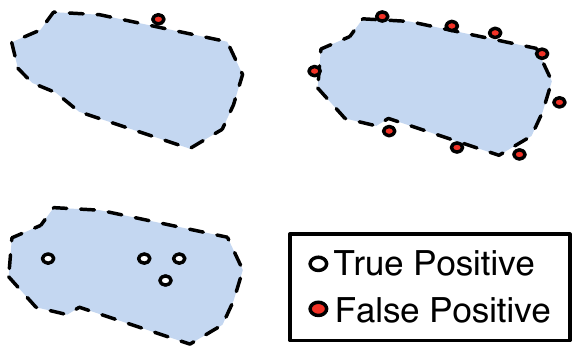}}\\
(a)&(b)&(c)&(d)
\end{tabular}
\end{center}
\caption{Plots of (a) an example master hypothesis $H$ of an image and (b) final unfiltered detections (yellow crosses) of an HoS ensemble with ground truth enclosed with red circles. The diagram in (c) illustrates three hypothetical pixel classifications of the same object, which highlights some drawbacks of pixel classification loss. The upper two classifications have equal loss and a tenth of the loss of the lower classification. In (d), the same object is detected by four different object detectors, highlighting that scoring object detectors is ill-posed.}
\label{f:azim}
\end{figure*}

\paragraph{Master detector} To finally detect objects in an image, the master hypothesis $H(\xv)$
is evaluated at every location $\xv$ in the image.
In Figure~\ref{f:azim}(a), {\em positive} master hypothesis objectness are
shown in green, and negative objectness in red. The intensity reflects
the magnitude of the cumulative objectness. 


There are numerous ways to extract the $(x, y)$ detections from the master
hypothesis, so it is necessary to compare some of them in an experiment.
Our baseline is to find large local maxima (LLM) of $H$, where
`large' is defined by a threshold.  Pre-smoothing the master hypothesis image
helps reduce multiple detections, and the LLM detector takes the pre-smoothing
radius as a parameter.  This parameter is trained on the validation set. Each point along the ROC curves in
Figure~\ref{f:roc} corresponds to a different thresholding of the LLM.

For comparison, we also employed a Kernel Density Estimate (KDE) detector. It
applies Kernel Density Estimation over the locations produced by the
(unsmoothed) LLM detector. The parameter for the KDE is the radius of the KDE
kernel.  The final detections are the large local maxima of the KDE, shown with
yellow crosshairs in Figure~\ref{f:azim}(b).

\input{experiments.tex}

%% file: experiments.tex
\subsection{Experiments}

\label{s:experiments}

Detecting small, convex objects in images is a natural application for
$(x,y)$ location-based object detection. We show that the HoS location
boosting algorithm performs very well compared to a much more
elaborate algorithm with many more parameters.  We compare HoS
boosting with the Beamer~\cite{eads2009bmvc} system on the `Arizona
dataset' described in that paper.  Beamer is an AdaBoost based system
using Grammar-guided Feature Extraction (GGFE), which relies on
elaborate post-processing of features and the master hypothesis and an
extensive, computationally expensive multidimensional grid search over
the post-processing parameters to maximize the AROC (Area under ROC)
score.

To facilitate a comparison between HoS boosting and Beamer we use
GGFE~\cite{eads2009bmvc} to generates a rich set of image features. We
use two grammars provided in the GGFE package, {\tt rich} and {\tt
haar}. The first grammar generates a broad set of non-linear image
features such as morphology, edge detection, Gabor filters, and
Haar-like~\cite{violajones01} features. The second grammar just
generates Haar-like features.

Beamer and HoS boosting use exactly the same training, validation, and testing
partitions as well as the same GGFE feature grammars. The parameters used
during training and validation are described in Table \ref{tab:params}.

In addition, we investigate the utility of the max operator in the loss
function by comparing the results of the HoS detector on the much simpler
smooth loss function.
Recall the non-smooth loss function:
\[
\LS =  \sum_{\xv \in \mbox{obj}} {\Exp{-H_t(\xv)}} + b \sum_{\xv \in \mbox{bg}} {\max \set{0, \Exp{H_t(\xv)}-1}}.
\]
We compare this to a smooth loss function which replaces the max with an
exponential,
\begin{equation}
\LS = \sum_{\xv \in \fg} \Exp{-H(\xv)} + b\sum_{x \in \bg}\Exp{H(\xv)}.
\end{equation}
The smooth function is considerably easier to optimize but is not able to ignore
large parts of the background.

\begin{table}
\begin{tabular}{l|l}
Parameter          & Value       \\\hline
$\delta$           &          10  pixels \\
$b$                &          $\frac{\mbox{num objects}}{\mbox{num bg pixels}}$  \\
Iterations         &          100  \\
Features per iteration         &          100  \\
$\rho$             &        7 pixels \\
{
Maximum false positive rate 
for AROC
}                  &        2.0 
\end{tabular}
\caption{Training parameters\label{tab:params}}
\end{table}

\paragraph{ROC curves and Validation}

To compare the detectors we use ROC curves which are truncated along the false 
positive rate axis. The reason for the truncation is that detectors achieving
a false positive rate exceeding two are of limited practical interest.

To generate a ROC curve, each prediction must be marked a true or
false positive, and the detection rate is simply the proportion of
objects found.  As the three examples in Figure~\ref{f:azim}(d) show,
scoring object detectors is ill-posed. A good detector for object
counting may be a poor detector for contour detection, tracking, or
target detection.  This motivates the need to choose a scoring metric
fits the object detection problem at hand. We chose to use the {\em
nearest neighbors metric} used by Beamer to evaluate car detection on
the Arizona data set. In this metric, true positives must be within
$\delta$ pixels of a ground truth object location. Additional
detections of the same object are false positives.  The parameters
giving the most favorable performance on the validation set are
applied to the final test set to give the reported results. During
validation for HoS boosting, Average Precision was used
as the validation criteria. For the competing detectors, the
area under the ROC curve was used to select the best parameters
during validation.

\begin{figure*}
\begin{center}
\begin{tabular}{cc}
\fbox{\includegraphics[width=7.5cm]{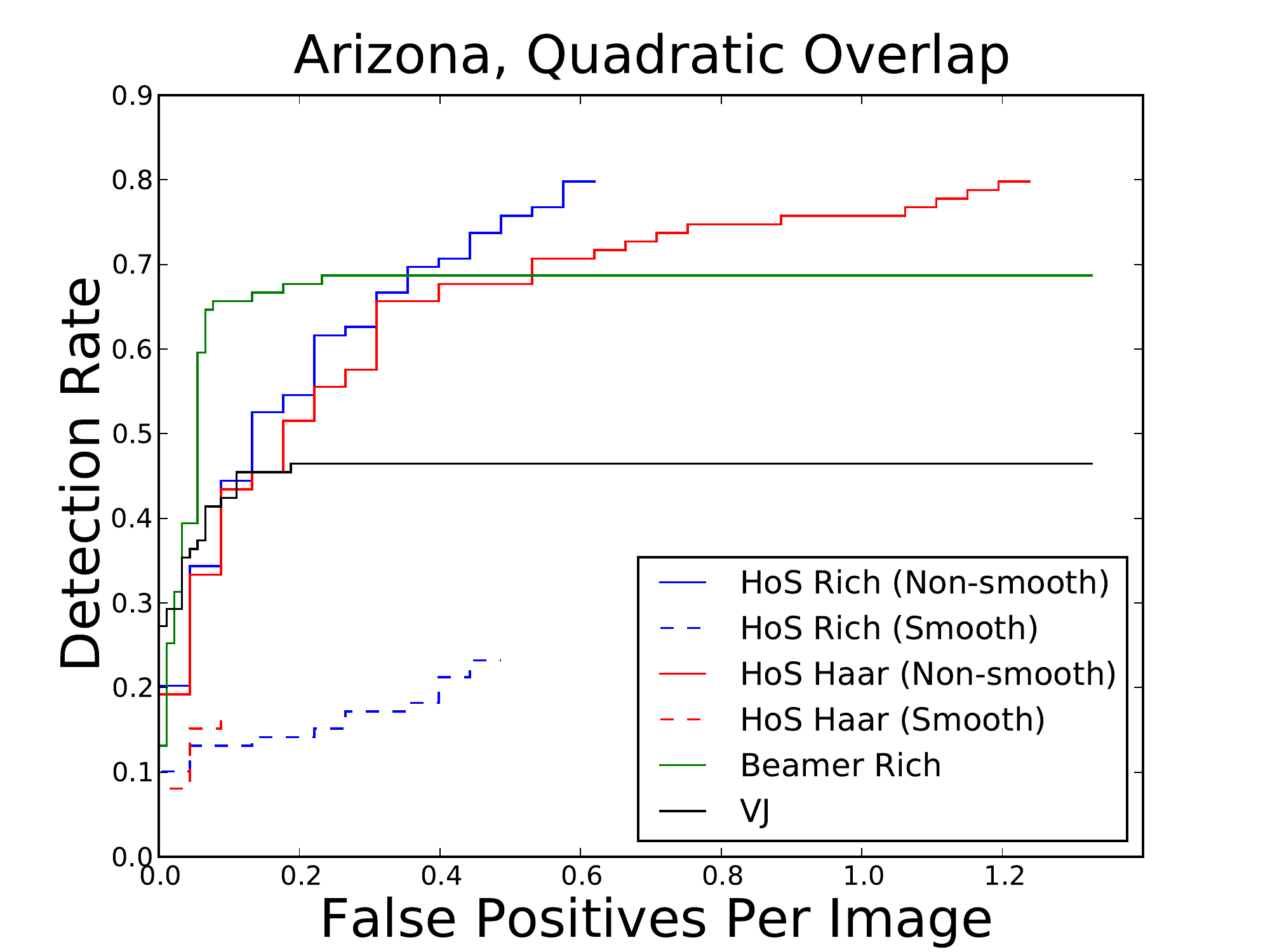}}&
\fbox{\includegraphics[width=7.5cm]{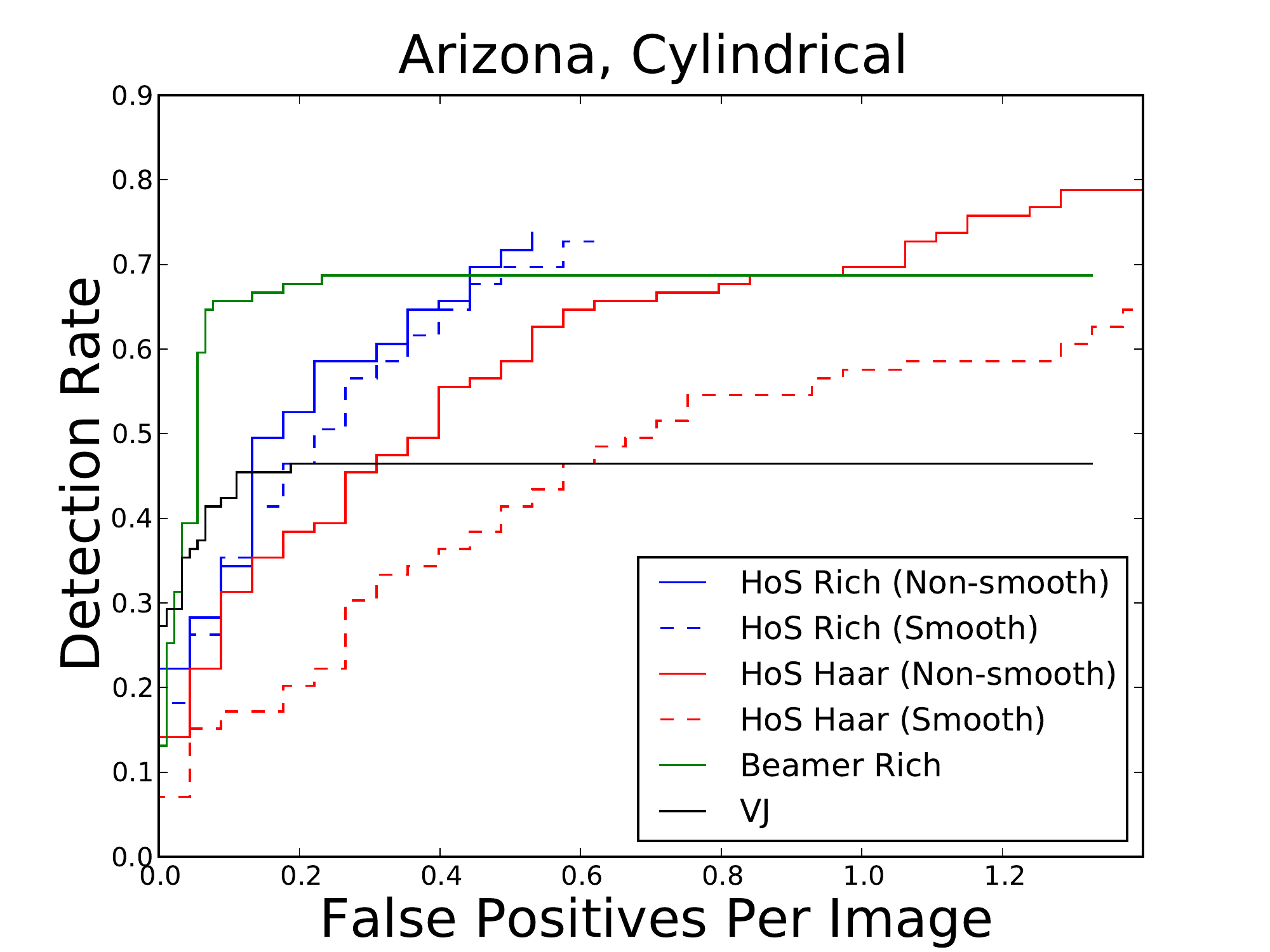}}\\
(a)&(b) \\
\end{tabular}
\caption{ Plot (a) shows the results with quadratic overlap correlation kernels and plot (b) shows results with cylindrical correlation kernels. ROC curves are presented for each competing detector applied to the test partition of the Arizona data set. These detectors include (a) location-boosted HoS ensembles, (b) Beamer, and (c) a variant of the Viola and Jones object detector. }
\label{f:roc}
\end{center}
\end{figure*}

\paragraph{Results Discussion}
Figure~\ref{f:roc} shows ROC curves on the testing partition of the
Arizona data set. HoS ensembles perform comparably to Beamer, a
competing pixel-based approach. Beamer's accuracy sharply increases
but quickly plateaus whereas the detection rate of the HoS ensemble
continues to rise. As shown in Figures~\ref{f:roc}(a-b), non-smooth
loss (solid) gives more favorable accuracy than smooth loss (dashed),
a rich set of features generated with a grammar outperforms using just
Haar features alone.

One interesting phenomenon worth noting is the ROC curves on
out-of-sample data sets appear to partially stabilize in fewer than
100 iterations. The detection rate of points at lower false positive
rates change minimally but the detection rate continues to improve at
higher false positive rates.  It is worth noting that HoS generalizes
on the Arizona data with significantly less optimization of training
parameters than Beamer.

%% file: conclusions.tex
\section{Conclusion}

We consider object detection in images where the desired objects 
are relatively rare and the bulk of the pixels are uninteresting background.
Pixel and widow-based methods must explicitly sift through the 
mass of dull background in order to find the desired objects.
In contrast, location-based approaches have the potential to 
zoom directly to the objects of interest.

We created a boosting formulation which uses weak hypotheses that predict a
list of confidence rated locations.  These are then filtered with the Hit or
Shift (HoS) filter to make weak detectors.  We used an exponential loss
reminiscent of AdaBoost, but modified so that it ignores background pixels
unless they are near a location predicted by a weak hypothesis in the ensemble.
Although our modified loss function is only piecewise differentiable, we
describe effective methods for optimizing the parameters and weights of the
weak hypothesis to minimize an upper bound on the loss.

We allow our weak hypotheses to have two different weights: one for the areas
near the predicted locations and a shift that reduces the objectness of
locations not near the predicted ones.  This shift works somewhat like negative
predictions and allows the master hypothesis to fix false positive predictions
by its ensemble members.  The structure of HoS allows us to optimize its three
parameters (the weight, the shift and threshold on detections) efficiently.
This is important because separately optimizing the shift and positive
predictions lets the master hypothesis make better use of marginal weak
hypothesis.  By finding the (approximately) optimal HoS parameters we can
search through a large number of features in each iteration of boosting.

Experimental results on a difficult data set show that our implementation gives
state-of-the-art performance, despite being considerably simpler and having
considerably fewer parameters than competing systems.  Since the method
proposed provides a framework for incorporating arbitrary weak object
detectors, we are confident that the technique has wide applicability.

Finally, with the permission of the original authors~\cite{eads2009bmvc}, we
will be making the small object data set available to others wishing to test
their algorithms.

